\documentclass[11pt]{article}

\PassOptionsToPackage{table,svgnames,dvipsnames}{xcolor}
\usepackage[preprint]{acl}

\usepackage{times}
\usepackage{latexsym}
\usepackage[T1]{fontenc}

\usepackage[utf8]{inputenc}

\usepackage{microtype}

\usepackage{inconsolata}

\usepackage{graphicx}

\usepackage{float}
\usepackage{booktabs}
\usepackage{svg}
\newcommand{\orcid}[1]{\href{https://orcid.org/#1}{\includesvg[width=10pt]{orcid}}}
\usepackage{mwe}
\usepackage{scrhack} 
\usepackage{xcolor}
\usepackage{listings}
\usepackage{lstautogobble}
\usepackage{tikz}
\usepackage{tabularx}
\usepackage{array}
\usepackage{amsmath}
\usepackage[dvipsnames]{xcolor}
\usepackage{multirow}

\definecolor{codegreen}{rgb}{0,0.6,0}
\definecolor{codegray}{rgb}{0.5,0.5,0.5}
\definecolor{codepurple}{rgb}{0.58,0,0.82}
\definecolor{backcolour}{rgb}{0.95,0.95,0.92}

\lstdefinestyle{mystyle}{
    backgroundcolor=\color{backcolour},   
    commentstyle=\color{codegreen},
    keywordstyle=\color{magenta},
    numberstyle=\tiny\color{codegray},
    stringstyle=\color{codepurple},
    basicstyle=\ttfamily\footnotesize,
    breakatwhitespace=false,         
    breaklines=true,                 
    captionpos=b,                    
    keepspaces=true,                 
    numbers=left,                    
    numbersep=5pt,                  
    showspaces=false,                
    showstringspaces=false,
    showtabs=false,                  
    tabsize=2
}

\title{Explaining News Bias Detection: A Comparative SHAP Analysis of Transformer Model Decision Mechanisms}

\author{
  \textbf{Himel Ghosh\textsuperscript{1,2}}
  \\
  \textsuperscript{1}Technical University of Munich, Germany,
  \textsuperscript{2}Sapienza University of Rome, Italy
  \\
  \small{
    \textbf{Correspondence:} \href{mailto:himel.ghosh@tum.de}{himel.ghosh@tum.de}
  }
}

\begin{document}
\maketitle
\begin{abstract}
Automated bias detection in news text is heavily used to support journalistic analysis and media accountability, yet little is known about how bias detection models arrive at their decisions or why they fail. In this work, we present a comparative interpretability study of two transformer-based bias detection models: a bias detector fine-tuned on the BABE dataset and a domain-adapted pre-trained RoBERTa model fine-tuned on the BABE dataset, using SHAP-based explanations. We analyze word-level attributions across correct and incorrect predictions to characterize how different model architectures operationalize linguistic bias. Our results show that although both models attend to similar categories of evaluative language, they differ substantially in how these signals are integrated into predictions. The bias detector model assigns stronger internal evidence to false positives than to true positives, indicating a misalignment between attribution strength and prediction correctness and contributing to systematic over-flagging of neutral journalistic content. In contrast, the domain-adaptive model exhibits attribution patterns that better align with prediction outcomes and produces 63\% fewer false positives. We further demonstrate that model errors arise from distinct linguistic mechanisms, with false positives driven by discourse-level ambiguity rather than explicit bias cues. These findings highlight the importance of interpretability-aware evaluation for bias detection systems and suggest that architectural and training choices critically affect both model reliability and deployment suitability in journalistic contexts.
\end{abstract}

\section{Introduction}
Detecting bias in news text is crucial to maintaining journalistic standards and helping readers assess media credibility. Natural language processing models trained to identify biased language in news articles promise to support these goals, but understanding how they make decisions is essential for trust and effective deployment \cite{bolukbasi2016mancomputerprogrammerwoman, Caliskan_2017}. As automated bias detection tools enter newsrooms and fact-checking workflows, transparency about their decision-making becomes critical \cite{blodgett-etal-2020-language}.

\begin{figure}[h]
    \centering
    \includegraphics[width=1\linewidth]{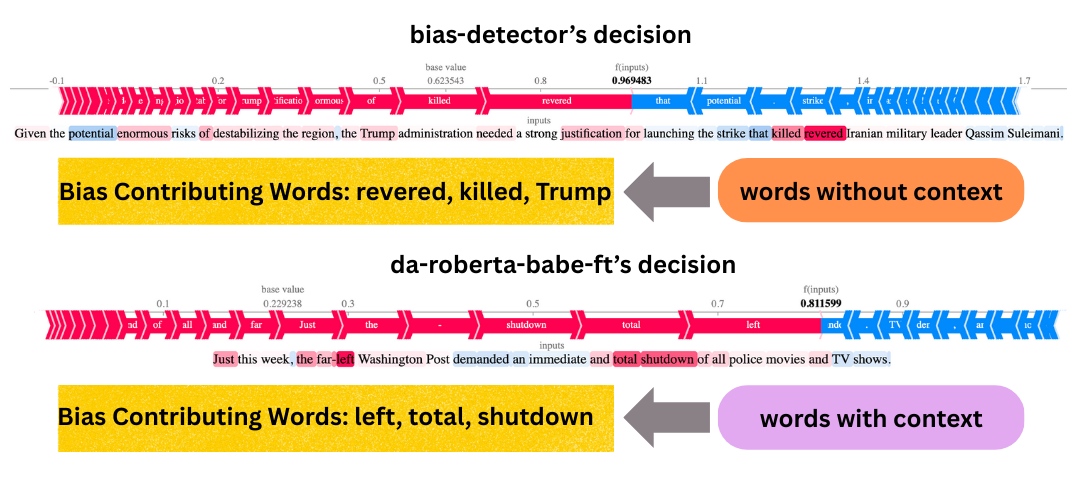}
    \caption{SHAP Reveals the Word Attributions for the model's decision. Red words increase bias, Blue words decrease bias in the sentence. This reveals, how the two models look differently into the sentence to make the decision.}
    \label{fig:shap_decisions}
\end{figure}

Model interpretability helps validate bias detection systems and diagnose failures. While accuracy metrics show overall performance, interpretability reveals which linguistic signals drive predictions that enable validation against journalistic principles and identification of problematic shortcuts \cite{lipton2017mythosmodelinterpretability, doshivelez2017rigorousscienceinterpretablemachine}. For journalism applications, interpretability is especially important: incorrect flagging can affect credibility, and stakeholders need to understand what constitutes bias from the model's perspective \cite{rudin2019stopexplainingblackbox}.

We apply SHAP (SHapley Additive exPlanations) to compare how two bias detection models: bias-detector \cite{ghosh2025} and DA-RoBERTa-BABE-FT \cite{Krieger2022} interpret news text on the BABE dataset \cite{spinde2021babe}. By analyzing word-level attributions across true positives, false positives, and true negatives, we investigate: (1) how models identify bias in news articles, (2) what linguistic patterns distinguish correct predictions from errors, and (3) how architectural differences affect interpretability and error patterns.
Our analysis reveals: (1) a reversal in attribution strength where false positives show higher SHAP magnitude than true positives in one model, indicating strong but misleading signals; (2) that models conflate emotional or evaluative language with social bias, suggesting limitations in contextual understanding relevant to journalism; and (3) significant differences between architectures in false positive rates and interpretability patterns, with DA-RoBERTa-BABE-FT showing 63\% fewer false positives and more intuitive SHAP alignments.

This work contributes: (1) a comparative interpretability framework using SHAP for bias detection models in news text; (2) a word-level attribution analysis revealing how different architectures operationalize bias in journalistic contexts; (3) identification of systematic misalignment between attribution strength and prediction correctness, particularly in false positive errors, and its implications for model reliability; and (4) insights into how bias detection models interpret linguistic signals, informing the development of more reliable systems for journalistic applications. Our code is available at this \href{https://drive.google.com/drive/folders/1gPEKmYq32v50WrMkMFKYjmHAgFBqOOXB?usp=sharing}{Link}.

\section{Related Work}
\subsection{Transformer Models for NLP}
Transformer architectures, introduced by \citet{vaswani}, have revolutionized NLP through their attention-based mechanism. \citet{devlin-etal-2019-bert} developed BERT, demonstrating the power of bidirectional pre-training for language understanding tasks. \citet{liu2019robertarobustlyoptimizedbert} introduced RoBERTa, a robustly optimized BERT variant that improved performance through better training procedures. These transformer-based models have become the foundation for many bias detection systems, including the Bias-Detector and DA-RoBERTa-BABE-FT model analyzed in this work. Understanding how these architectures process and interpret linguistic signals is essential for developing more reliable bias detection systems.
\subsection{Bias Detection in News}
Automated bias detection in news text has gained attention with the rise of computational journalism. \citet{spinde2021babe} introduced BABE (Bias Analysis by Experts), a sentence-level dataset for bias detection in news articles using distant supervision from expert annotations. Their work demonstrated the feasibility of neural models for identifying biased language in journalistic contexts. Building on this foundation, \citet{ghosh2025} developed the bias-detector model, which we analyze in this work, focusing specifically on detecting bias in news articles. \citet{Krieger2022} advanced the field by proposing domain-adaptive pre-training approaches for language bias detection in news, showing that specialized pre-training improves performance on media bias tasks.
Beyond news-specific applications, research on bias detection has expanded to measure and identify various forms of social bias in language models. \citet{nadeem-etal-2021-stereoset} introduced StereoSet, a benchmark for measuring stereotypical bias in pretrained language models, while \citet{nangia-etal-2020-crows} created CrowS-Pairs, a challenge dataset for measuring social biases in masked language models. \citet{powers2025} proposed the GUS Framework for benchmarking social bias classification using both discriminative and generative language models, providing a comprehensive evaluation framework. \citet{hutchinson-etal-2020-social} examined social biases in NLP models as barriers for persons with disabilities, highlighting the importance of inclusive bias detection. \citet{borkan2019nuancedmetricsmeasuringunintended} developed nuanced metrics for measuring unintended bias with real data for text classification, while \citet{Rottger_2021} introduced HateCheck, a functional test suite for hate speech detection models. These works collectively establish the landscape of bias detection and measurement in NLP, with our work focusing specifically on interpretability of bias detection models in news text.
\subsection{Interpretability Methods in NLP}
Understanding how NLP models make decisions has become increasingly important as these systems are deployed in high-stakes applications. Interpretability methods can be broadly categorized into model-agnostic approaches that work with any model and model-specific approaches that leverage internal model representations. \citet{ribeiro2016whyitrustyou} introduced LIME (Local Interpretable Model-agnostic Explanations), a model-agnostic method that explains individual predictions by approximating the model locally with interpretable models. \citet{lundberg2017unifiedapproachinterpretingmodel} proposed SHAP (SHapley Additive exPlanations), which unifies multiple explanation methods under a unified framework based on Shapley values from cooperative game theory, providing theoretically grounded feature attributions.
The interpretability of attention mechanisms in transformer models has been particularly debated. \citet{jain2019attentionexplanation} argued that attention weights do not necessarily provide faithful explanations, challenging the common assumption that attention can be directly interpreted. \citet{wiegreffe2019attentionexplanation} responded by demonstrating cases where attention can indeed provide meaningful explanations, while \citet{serrano2019attentioninterpretable} provided a more nuanced analysis of when attention can be considered interpretable. These debates highlight the complexity of interpreting neural models and the need for rigorous evaluation of interpretability methods. \citet{deyoung2020eraserbenchmarkevaluaterationalized} introduced ERASER, a benchmark for evaluating rationalized NLP models, providing standardized metrics for assessing explanation quality. In this work, we employ SHAP as our primary interpretability method, leveraging its theoretical foundations and model-agnostic nature to compare different bias detection architectures.
\subsection{SHAP Applications in NLP}
SHAP has been widely adopted for explaining NLP models due to its theoretical grounding and ability to provide consistent, locally accurate explanations. \citet{lundberg2019explainableaitreeslocal} extended SHAP to tree-based models, demonstrating how local explanations can be aggregated to achieve global understanding. \citet{Mosca2022SHAPBasedEM} provided a comprehensive review of SHAP-based explanation methods specifically for NLP interpretability, discussing applications across various NLP tasks and highlighting best practices. \citet{covert2022explainingremovingunifiedframework} developed a unified framework for model explanation based on the principle of explaining by removing features, which aligns with SHAP's approach of measuring feature contributions through marginalization.
Despite the widespread use of SHAP in NLP, relatively little work has applied it to understand bias detection models specifically, particularly in comparative settings. Most existing applications focus on single-model explanations or task-specific interpretability. Our work extends this line of research by applying SHAP to conduct a comparative analysis of two different bias detection architectures, revealing how model design choices affect interpretability patterns and error modes in news bias detection.
\subsection{Model Comparison Methods}
\citet{Dietterich} proposed statistical tests for comparing supervised classification algorithms, emphasizing the importance of accounting for variance in performance estimates. \citet{dror} provided a comprehensive guide to testing statistical significance in natural language processing, highlighting common pitfalls and best practices in model comparison. While these frameworks are important for rigorous evaluation, our work focuses on interpretability-based comparison rather than purely statistical comparisons, examining how different models interpret the same inputs rather than just comparing their performance metrics.

\section{Methodology}

\subsection{Dataset and Models}
We evaluate two bias detection models on the BABE (Bias Analysis by Experts) dataset \cite{spinde2021babe}, a sentence-level dataset for bias detection in news articles. BABE contains expert-annotated sentences from news sources, labeled as biased or non-biased, making it suitable for analyzing how models interpret linguistic signals in journalistic contexts.

\textbf{Models:} We analyze two pre-trained state-of-the-art bias detection models: (1) bias-detector\footnote{\url{https://huggingface.co/himel7/bias-detector}} \cite{ghosh2025}, a model specifically designed for detecting bias in news text, and (2) DA-RoBERTa-BABE-FT\footnote{\url{https://huggingface.co/mediabiasgroup/DA-RoBERTa-BABE-FT}}, a domain-adaptive RoBERTa model \cite{liu2019robertarobustlyoptimizedbert} fine-tuned on the BABE dataset \cite{Krieger2022}. Both models are transformer-based architectures \cite{vaswani, devlin-etal-2019-bert} that classify sentences as biased or non-biased. We use the test split of BABE for evaluation, ensuring that our analysis reflects model behavior on held-out data.

\textbf{Preprocessing:} For each model, we use the corresponding tokenizer from the HuggingFace Transformers library. Input sequences are tokenized with a maximum length of 256 tokens, following standard practices for sentence classification tasks. Labels are normalized to binary format (0 for non-biased, 1 for biased) to ensure consistent analysis across models.
\subsection{SHAP-Based Interpretability Framework}
We employ SHAP (SHapley Additive exPlanations) \cite{lundberg2017unifiedapproachinterpretingmodel} to explain model predictions. SHAP is based on Shapley values from cooperative game theory \cite{shapley1953}, which provide a theoretically grounded method for attributing prediction contributions to individual features.

\textbf{Theoretical Foundation:} Shapley values distribute the total prediction value among features according to their marginal contributions across all possible feature coalitions. For a model prediction $f(x)$ on input $x$ with features $F$, the Shapley value $\phi_i$ for feature $i$ is:
\begin{equation}
\phi_i = \sum_{S \subseteq F \setminus \{i\}} \frac{|S|!(|F|-|S|-1)!}{|F|!} [f(S \cup \{i\}) - f(S)]
\end{equation}
where $S$ represents a subset of features, and $f(S)$ denotes the model output when only features in $S$ are present. This formulation ensures that attributions satisfy desirable properties: efficiency (attributions sum to the prediction difference from baseline), symmetry (symmetric features receive equal attributions), dummy (features with no effect receive zero attribution), and additivity (attributions are additive across model outputs).

\textbf{SHAP Implementation:} We use the SHAP library's TextExplainer with a Text masker, which handles tokenization and masking for transformer models. For each input sentence, SHAP generates token-level attributions by computing Shapley values for each token's contribution to the model's prediction. The explainer uses a background dataset (a subset of the test set) to establish baseline expectations, and computes attributions by marginalizing over token presence/absence.

\textbf{Prediction Function:} For each model, we construct a prediction function that takes a list of text inputs and returns class probabilities. The function tokenizes inputs, passes them through the model, and applies softmax to obtain probability distributions over classes. SHAP then explains these probability outputs, providing attributions for the positive (biased) class probability, which we use as our primary analysis target.

\subsection{Word-Level Attribution Aggregation}

\begin{figure}[h]
    \centering
    \includegraphics[width=1\linewidth]{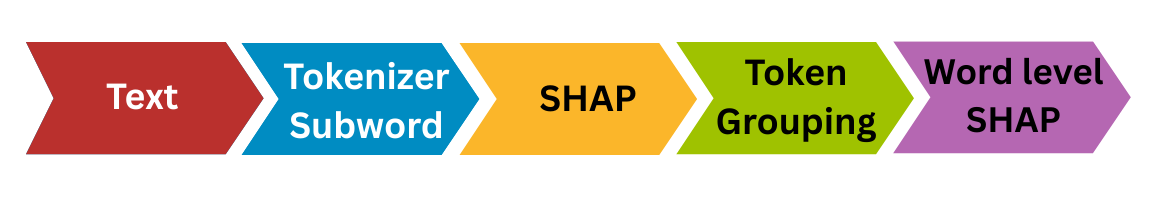}
    \caption{Word-level SHAP aggregation pipeline. Token-level SHAP values produced by a transformer model are grouped into words based on tokenizer boundaries and text alignment, then aggregated to produce interpretable word-level attributions.}
    \label{fig:shap_pipeline}
\end{figure}

Transformer-based models operate on subword units, while bias analysis is more interpretable at the word level. To bridge this gap, we aggregate token-level SHAP attributions into word-level explanations using a structured aggregation pipeline, illustrated in Figure~\ref{fig:shap_pipeline}.

Given an input sentence, we first compute token-level SHAP values using the model-specific tokenizer and SHAP TextExplainer. Subword tokens are then grouped into words based on tokenizer boundary markers and original text alignment. For each word, we aggregate the SHAP values of all constituent tokens by summation, preserving the additive property of Shapley values. This produces a single attribution score per word, reflecting its overall contribution to the model’s predicted probability of the biased class.

To ensure linguistic consistency, we apply normalization steps that correct tokenization artifacts and align word boundaries with the original text. These steps prevent spurious attributions arising from subword fragmentation or punctuation handling. The resulting word-level attributions enable consistent comparison across instances and models while retaining faithfulness to the underlying token-level explanations.

\subsection{Stratified Sampling Strategy} To ensure comprehensive analysis across different prediction types, we employ stratified sampling to select instances for SHAP analysis. For each model, we first evaluate all test instances to obtain predictions and ground truth labels. We then categorize instances into: (1) True Positives (TP): correctly predicted as biased, (2) False Positives (FP): incorrectly predicted as biased, (3) True Negatives (TN): correctly predicted as non-biased, and (4) False Negatives (FN): incorrectly predicted as non-biased. 

\textbf{Sampling Procedure:} We sample up to 100 instances from each category (TP, FP, TN) to ensure balanced representation. For the bias-detector model, we analyze 237 instances total (100 TP, 37 FP, 100 TN), while for DA-RoBERTa-BABE-FT, we analyze 212 instances (100 TP, 12 FP, 100 TN). The sampling is performed with a fixed random seed to ensure reproducibility. This stratified approach helps us to compare SHAP patterns across prediction categories and identify systematic differences between correct and incorrect predictions.

\subsection{Analysis Framework}
\textbf{Per-Instance Analysis:} For each sampled instance, we compute word-level SHAP attributions and store: (1) the original text, (2) ground truth label, (3) predicted label, (4) predicted probability for the biased class, (5) token-level SHAP values, and (6) aggregated word-level attributions with word frequencies.

\textbf{Global Word Importance:} We aggregate word-level attributions across all instances to compute global word importance metrics. For each unique word, we compute: (1) mean absolute SHAP value across all occurrences, (2) total occurrence count, and (3) mean signed SHAP value (indicating whether the word generally increases or decreases bias probability). This global analysis reveals which words are most influential across the dataset and how they contribute to model predictions.

\textbf{Category-Specific Analysis:} We analyze SHAP patterns separately for TP, FP, and TN instances to understand how model interpretations differ between correct and incorrect predictions. For each category, we compute: (1) mean absolute SHAP value per instance (indicating overall attribution magnitude), (2) top words by mean absolute SHAP value, (3) distribution of positive vs. negative SHAP values, and (4) word frequency patterns. This category-specific analysis enables identification of patterns that distinguish correct predictions from errors.

\textbf{Comparative Analysis:} We compare SHAP patterns between the two models to understand how architectural differences affect interpretability. We examine: (1) differences in top bias indicators, (2) variations in false positive patterns, (3) differences in SHAP magnitude distributions across prediction categories, and (4) model-specific word interpretation patterns.

\subsection{Statistical Significance Testing}
To assess whether observed differences between models are statistically significant, we employ McNemar's test \cite{Dietterich}. McNemar's test is appropriate for comparing two classifiers on the same test set, as it accounts for the paired nature of predictions (both models make predictions on identical instances).

\textbf{McNemar's Test:} We construct a 2×2 contingency table where: (a) both models are correct, (b) Model 1 is wrong and Model 2 is correct, (c) Model 1 is correct and Model 2 is wrong, and (d) both models are wrong. The test statistic is computed as:
\begin{equation}
    \chi^2 = \frac{(|b - c| - 1)^2}{b + c}
\end{equation}
with 1 degree of freedom, where $b$ and $c$ represent the off-diagonal counts. We apply continuity correction for small sample sizes. The null hypothesis is that the two models have the same error rate (i.e., $b = c$). A significant result indicates that the models differ in their prediction patterns, which may reflect differences in how they interpret linguistic signals.

\subsection{Evaluation Metrics}
We report standard classification metrics: accuracy, precision, recall, and F1 score (both binary and macro-averaged). Additionally, we analyze SHAP-specific metrics: (1) mean absolute SHAP value per instance category (TP, FP, TN), (2) distribution of SHAP values across words, and (3) word-level attribution patterns. These metrics provide both performance and interpretability perspectives on model behavior.

\section{Evaluation Results}
\subsection{Overall Model Performance}
Table~\ref{tab:performance} summarizes the performance metrics for both models.
While bias-detector achieved higher accuracy and F1, DA-RoBERTa-BABE-FT showed better precision. The significant difference in false positive rates (15.6\% vs. 5.7\%) suggests that architectural differences and training approaches substantially affect precision in bias detection tasks.

\begin{table}[htbp]
\centering
\caption{Model Performance on BABE Test Set}
\label{tab:performance}
\resizebox{\linewidth}{!}{
\begin{tabular}{lcc}
\toprule
\textbf{Metric} & \textbf{bias-detector} & \textbf{da-roberta-babe-ft}\\
\midrule
Accuracy & 0.8520 & 0.8230 \\
Precision & 0.9237 & 0.9704 \\
Recall & 0.8014 & 0.7048 \\
Binary F1 & 0.8582 & 0.8166 \\
\rowcolor{gray!10}
\textbf{Macro F1} & \textbf{0.8517} & 0.8228 \\
Weighted F1 & 0.8525 & 0.8221 \\
\bottomrule
\end{tabular}
}
\end{table}

\subsection{SHAP Magnitude Patterns}
\begin{figure}[h]
    \centering
    \includegraphics[width=1\linewidth]{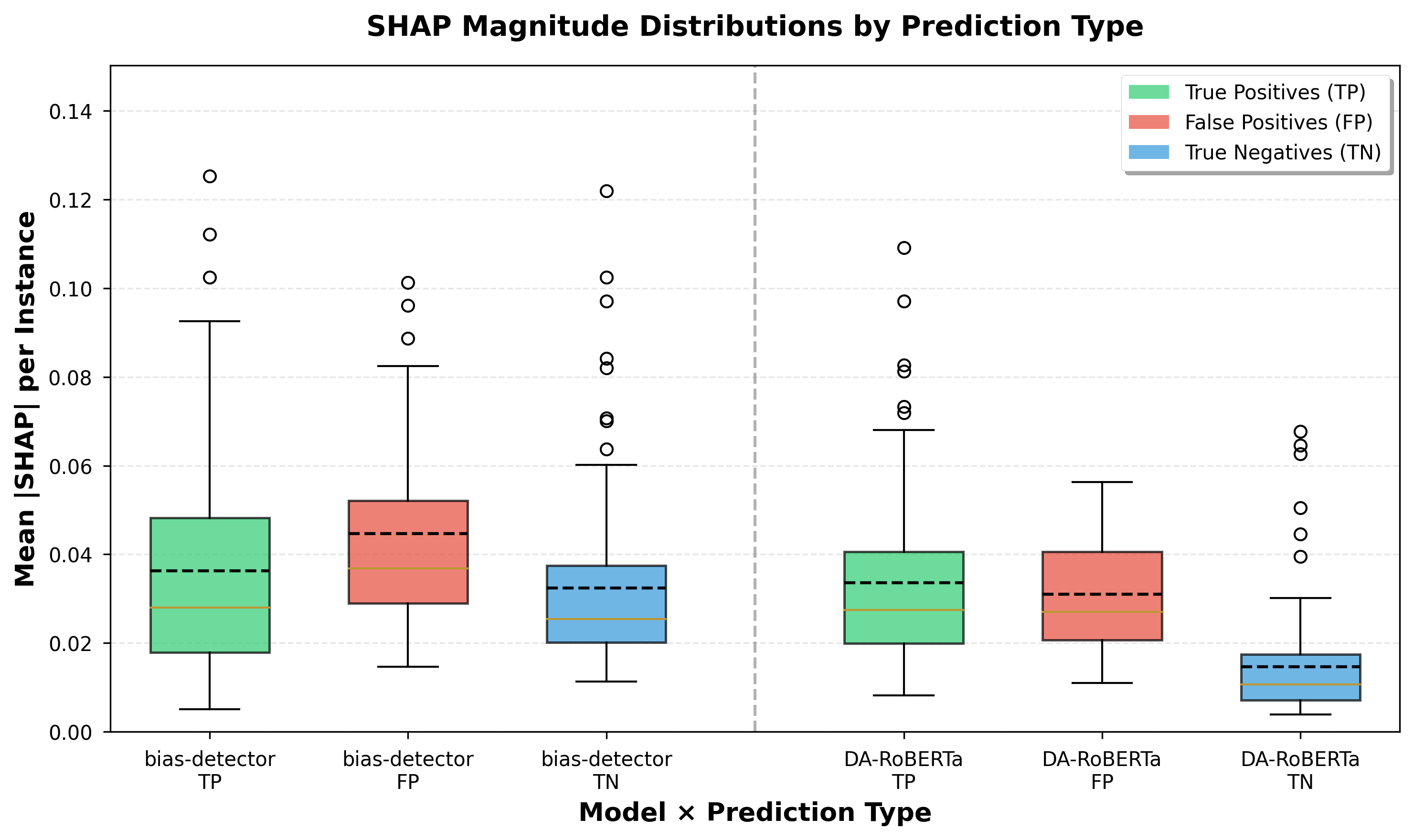}
    \caption{SHAP Magnitude Distributions: False positives in bias-detector exhibit higher attribution magnitude than true positives, while DA-RoBERTa-BABE-FT shows the opposite, indicating better alignment between explanations and predictions.}
    \label{fig:smd}
\end{figure}

Word-level SHAP attribution magnitudes reveal differences in how models interpret linguistic signals. Fig.~\ref{fig:smd} presents mean absolute SHAP values by prediction category for both models.

This pattern of bias-detector indicates that incorrect predictions (false positives) exhibit stronger individual word signals than correct predictions, suggesting a a reversal in attribution strength between false and true positives where the model relies on strong but misleading linguistic cues. However, DA-RoBERTa-BABE-FT model exhibited the opposite pattern where correct predictions show stronger signals than errors suggesting that its SHAP attributions better align with actual model decision-making. DA-RoBERTa-BABE-FT showed lower confidence on false positives, indicating less overconfidence on incorrect predictions. This suggests DA-RoBERTa-BABE-FT is less overconfident on incorrect predictions.

\subsection{Global Bias Indicators}
\begin{figure}[h]
    \centering
    \includegraphics[width=1\linewidth]{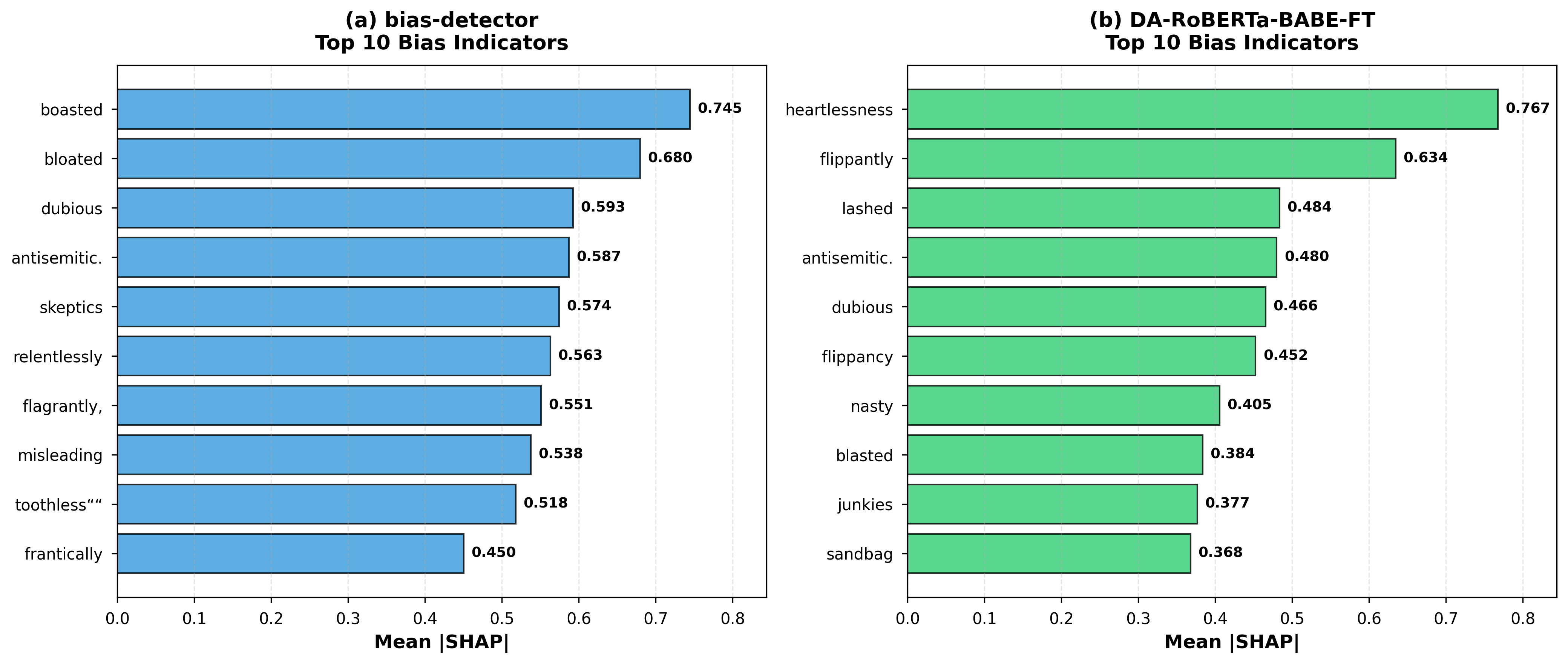}
    \caption{Top Bias Indicators for the bias-detector model.}
    \label{fig:tbi}
\end{figure}

We aggregated word-level SHAP attributions across all instances to identify global bias indicators. Fig.~\ref{fig:tbi} presents the top 10 words with highest mean absolute SHAP values for each model.

For bias-detector, Top indicators include emotional and evaluative language: "boasted", "bloated", "dubious", "antisemitic", and "skeptics". Many top words are negative descriptors ("bloated", "dubious") or explicit bias terms ("antisemitic"), suggesting the model relies heavily on loaded language and explicit bias markers. While for DA-RoBERTa-BABE-FT, top indicators emphasize dismissive and moral language: "heartlessness", "flippantly", "lashed", "antisemitic", and "dubious". This model shows a stronger focus on dismissive language patterns ("flippantly", "heartlessness") and emotional descriptors compared to the original model. Some overlap exists ("dubious", "antisemitic", "lashed"), suggesting consensus on certain linguistic bias markers.

\subsection{Word-Level Patterns}

\begin{figure}[h]
    \centering
    \includegraphics[width=1\linewidth]{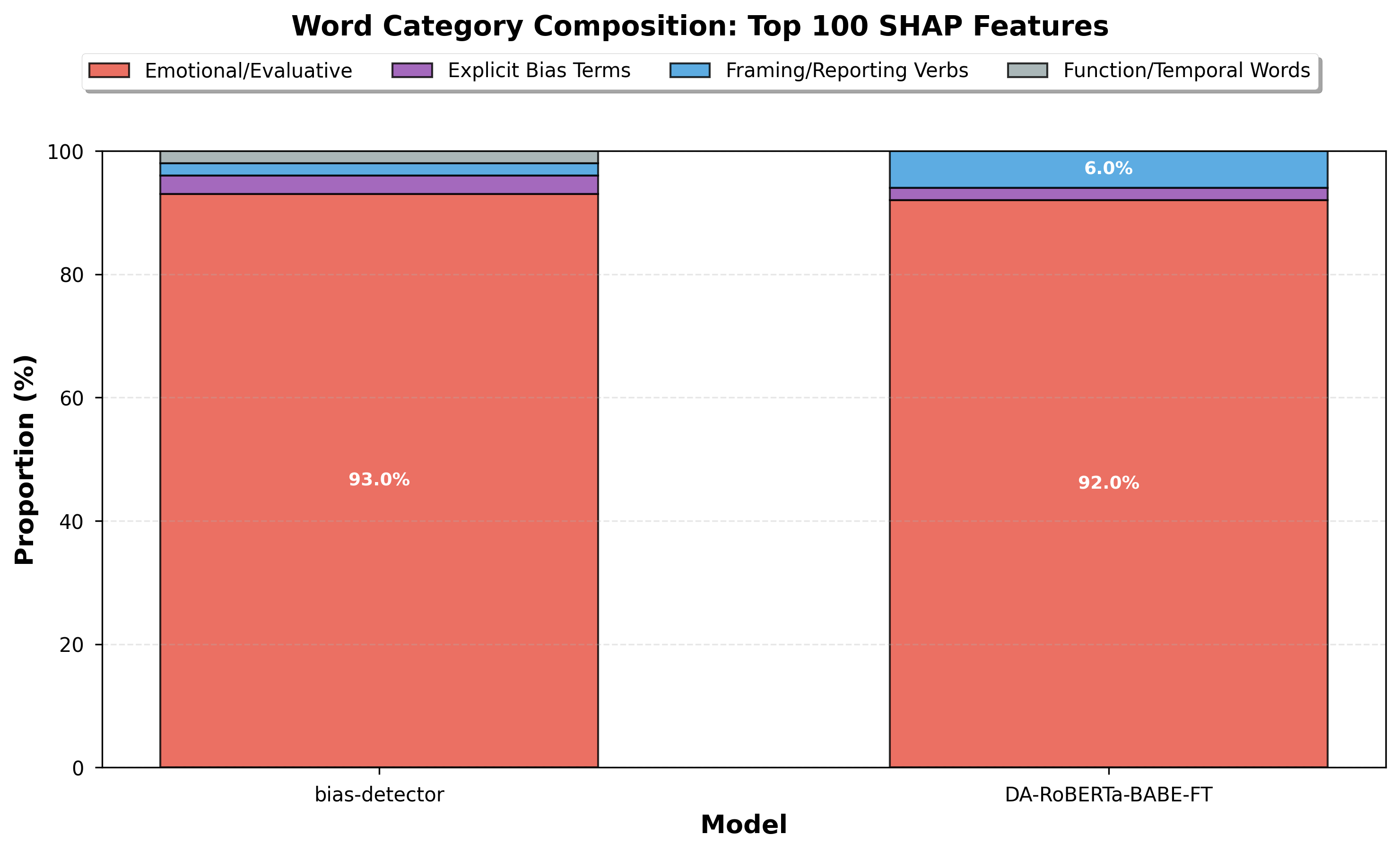}
    \caption{Word category composition of the top 100 SHAP-attributed features for each model.}
    \label{fig:wcc}
\end{figure}

Figure \ref{fig:wcc} presents the lexical category composition of the top 100 most influential SHAP features for both models. For both the bias-detector and DA-RoBERTa-BABE-FT, evaluative and emotionally charged language dominates the highest-attribution features, reflecting the fact that bias across political, social, and cultural domains is often linguistically realized through subjective or judgment-laden expressions. At this aggregate level, the two models appear superficially similar in terms of the lexical categories they attend to.

However, this similarity conceals important differences in how such features are operationalized during prediction. As shown by earlier attribution magnitude analyses, the bias-detector tends to treat evaluative language as sufficient evidence for bias, often producing high-confidence predictions even in neutral reporting contexts. In contrast, DA-RoBERTa-BABE-FT assigns relatively greater importance to framing and reporting verbs, suggesting that evaluative expressions are more frequently interpreted in relation to attribution, quotation, or narrative context rather than as standalone bias indicators.

\begin{figure}[h]
    \centering
    \includegraphics[width=1\linewidth]{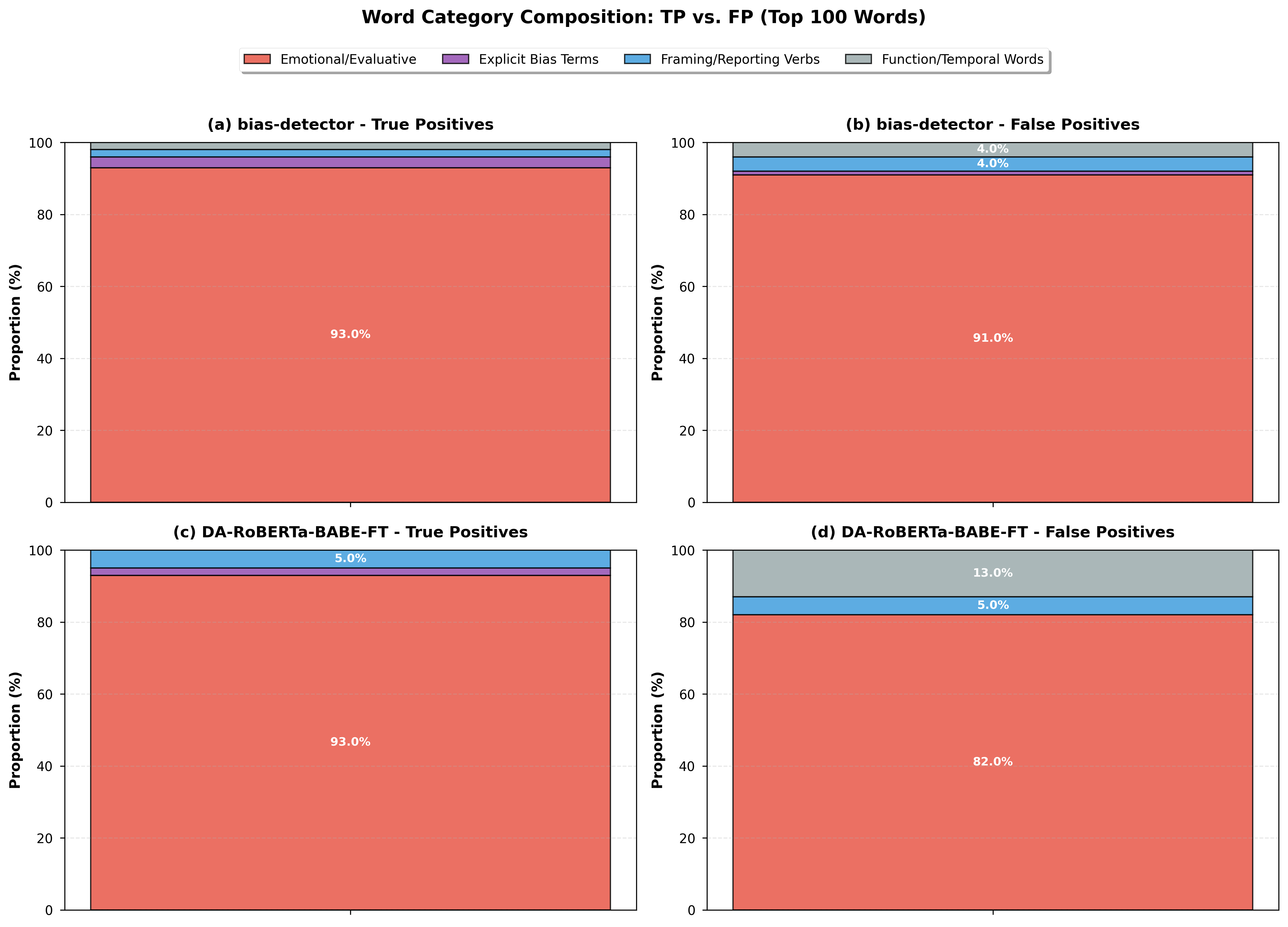}
    \caption{Word category composition of the top 100 SHAP-attributed tokens for true positive and false positive predictions, separated by model.}
    \label{fig:wcctpfp}
\end{figure}

Figure \ref{fig:wcctpfp} further disambiguates these behaviors by conditioning the analysis on prediction correctness. For the bias-detector, the category distribution remains nearly unchanged between true positives and false positives, with evaluative language dominating both. This invariance indicates that the model relies on similar lexical cues regardless of whether its prediction is correct, explaining its tendency toward systematic false positives driven by emotional tone rather than contextual bias.

By contrast, DA-RoBERTa-BABE-FT exhibits a pronounced shift between correct and incorrect predictions. While true positives are primarily driven by evaluative language, false positives show substantially increased reliance on function and temporal words, alongside weaker evaluative signals. This pattern suggests that DA-RoBERTa-BABE-FT’s errors arise in structurally or discourse-level ambiguous cases where explicit semantic bias cues are limited, rather than from indiscriminate overgeneralization of emotional language. Taken together, these findings indicate that although both models attend to similar lexical categories, they differ fundamentally in how those features are integrated into bias judgments.

\subsection{Statistical Significance}
We applied McNemar's test to assess whether the observed difference in error patterns between the two models is statistically significant. On the test set, we constructed a 2$\times$2 contingency table comparing correct and incorrect predictions for both models. McNemar's test yielded $\chi^2$ = 5.723 with p = 0.0167, indicating a statistically significant difference in model errors. This result suggests that the difference in false positive rates (15.6\% vs. 5.7\%) is unlikely to be due to chance, supporting the conclusion that DA-RoBERTa-BABE-FT achieves higher precision, while the bias-detector attains higher overall F1 score.

\section{Conclusion}

In this work, we conducted a comparative interpretability analysis of two transformer-based bias detection models for news text using SHAP-based explanations. While both models achieve competitive classification performance, the bias-detector assigns stronger internal evidence to false positive predictions than to correct detections, indicating a misalignment between attribution strength and prediction correctness. Such behavior leads to systematic over-flagging of neutral journalistic content, raising concerns for deployment in real-world editorial settings.

By contrast, DA-RoBERTa-BABE-FT produces significantly fewer false positives and exhibits attribution patterns that align more closely with prediction outcomes, with stronger evidence supporting correct detections and weaker, more distributed signals associated with errors. Although both models rely heavily on evaluative language, they operationalize bias differently: the bias-detector emphasizes explicit negatively valenced descriptors, whereas DA-RoBERTa-BABE-FT more consistently contextualizes evaluative expressions through framing and attribution patterns.

Overall, our results highlight the importance of interpretability-aware evaluation when developing and deploying bias detection models for journalistic applications. Beyond raw accuracy, understanding how and why models make decisions is essential for ensuring responsible use, minimizing false positives, and maintaining trust in automated media analysis tools. Future work should explore context-sensitive modeling approaches and calibration-aware training strategies to further improve the robustness of bias detection systems.


\bibliography{custom}

\section{Ethics and Broader Impact}

Automated bias detection systems have the potential to support journalism, media analysis, and public accountability by highlighting subjective or biased language at scale. However, such systems also carry ethical risks, particularly when deployed in sensitive editorial or political contexts. Incorrect bias flagging may undermine journalistic credibility, mischaracterize neutral reporting, or disproportionately affect coverage of contentious topics. Our findings demonstrate that false positives remain a central challenge, underscoring the need for careful evaluation before real-world deployment.

We position these systems as assistive tools whose outputs should be interpreted alongside human judgment. By focusing on interpretability through SHAP-based analysis, our study aims to improve transparency and enable practitioners to better understand model behavior, identify systematic failure modes, and assess the reliability of predictions. Such transparency is particularly important given the context-dependent nature of bias and the diversity of journalistic norms across cultures and outlets.

The models evaluated in this study are trained on the BABE dataset, which reflects expert annotations and inherent normative assumptions about bias. As a result, model predictions and their explanations may encode annotation biases or dataset-specific perspectives. Our analysis highlights how architectural and training choices influence not only performance but also the types of linguistic signals models prioritize, which may affect downstream fairness and generalization.

We emphasize that interpretability techniques, including SHAP, do not eliminate bias or guarantee correct model behavior. Rather, they provide a diagnostic lens for examining model decisions and guiding more responsible development. Future work should consider broader datasets, cross-cultural bias definitions, and human-in-the-loop evaluation frameworks to mitigate potential harms and support ethical deployment of bias detection systems.

\appendix
\section{Appendix: Word-Level Attribution Aggregation in Detail}
Transformer models operate on subword tokens (e.g., BPE tokens in RoBERTa), but interpretability for bias detection benefits from word-level analysis. We aggregate token-level SHAP values to word-level attributions through a multi-step process.

\textbf{Token-to-Word Mapping:} We identify word boundaries using tokenization markers. RoBERTa-style tokenizers use special markers (e.g., Ġ or $\_$) to indicate tokens that begin new words. We detect these markers and group consecutive subword tokens that belong to the same word. For tokens without explicit markers, we use whitespace information from the original text to determine word boundaries.

\textbf{Word Boundary Detection:} To handle cases where tokens may be incorrectly merged (e.g., "tuesdayquoting" or "nationalismfueled"), we implement heuristic word boundary detection. We split tokens when: (1) a lowercase letter is followed by an uppercase letter (e.g., "tuesdayQuoting" → ["tuesday", "Quoting"]), (2) common word endings (e.g., "day", "ing", "ed", "ly") are followed by lowercase letters that could start new words, or (3) punctuation or hyphens indicate word separation.

\textbf{Attribution Aggregation:} For each word, we sum the SHAP values of all tokens that comprise that word. If a word is split across multiple tokens, the word-level attribution is the sum of all constituent token attributions. This aggregation preserves the additive property of SHAP values while providing interpretable word-level explanations.

\textbf{Normalization:} We apply normalization to handle tokenization artefacts. Merged words (e.g., "dmnboasted" → "boasted") are split, and punctuation artifacts are stripped from word boundaries. This ensures that our word-level analysis reflects actual linguistic patterns rather than tokenization idiosyncrasies.

\section{Appendix: Token Level SHAP Examples}
\subsubsection{True Positives} 
Refer to Fig. \ref{fig:appendixA} below.
\begin{figure}[h]
    \centering
    \includegraphics[width=1\linewidth]{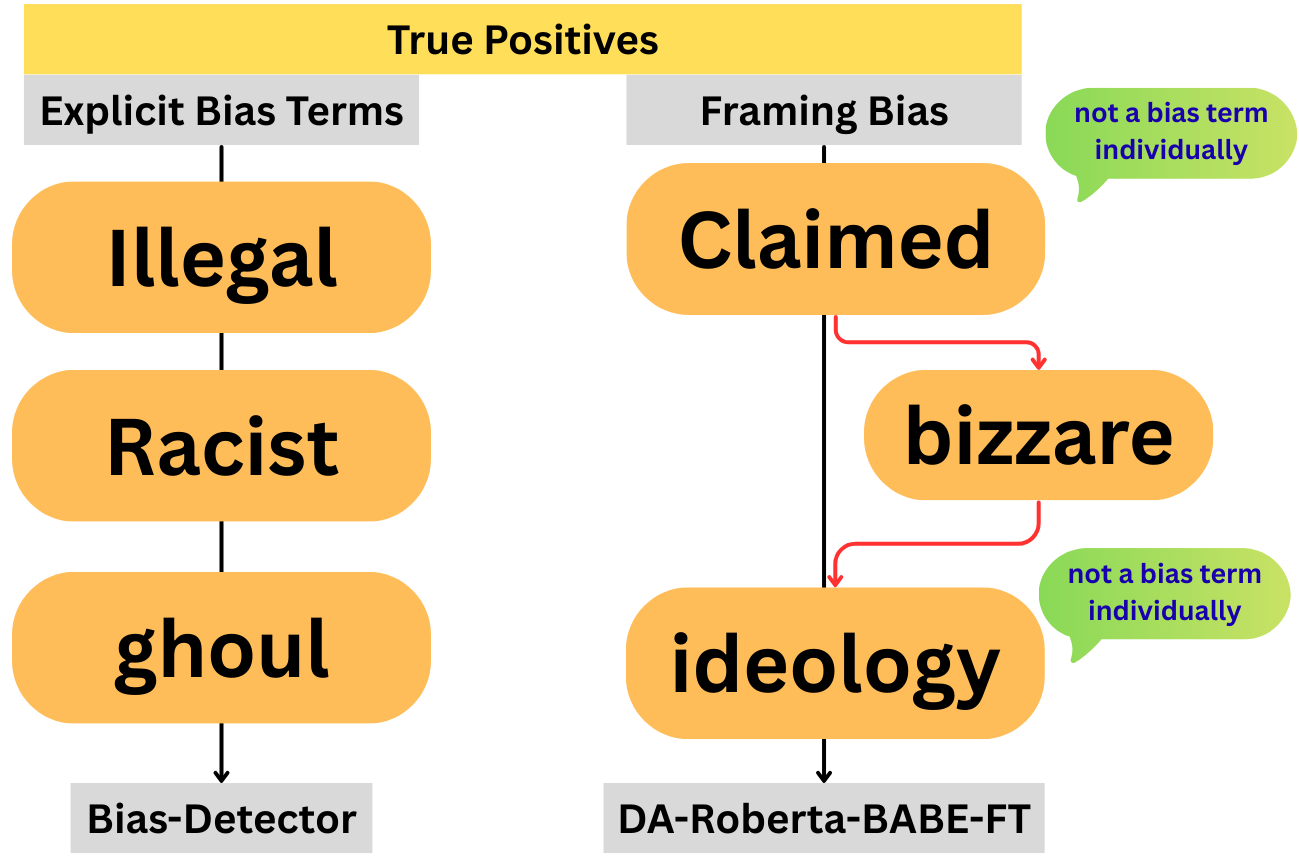}
    \caption{Examples of Biased Terms detected by both the models for the True Positives.}
    \label{fig:appendixA}
\end{figure}

\subsubsection{False Positives}
Bias-Detector: False positives exhibit the highest SHAP magnitude (0.0354) despite being incorrect predictions. Top words include: "illegal" (0.1400, positive), "said" (0.1304, negative), "tuesday" (0.1128, negative), and "could" (0.1094, negative). Crucially, "illegal" appears in both true positives (0.3489) and false positives (0.1400), indicating context-dependent interpretation where the same word can signal bias or not depending on surrounding context.
Many top false positive words have negative SHAP values (indicating they should decrease bias probability), yet the model predicts bias with high confidence (0.796). This suggests non-linear interactions or missing contextual features that word-level SHAP cannot capture. Function and temporal words dominate false positives, with intensifiers ("highly": 0.0624, "very": 0.0599) showing positive signals, indicating the model may conflate emotional intensity with bias.

DA-RoBERTa-BABE-FT: False positives show lower SHAP magnitude (0.0215) and lower confidence (0.673), indicating more reasonable error patterns. Top words include: "claims" (0.2061, positive), function words ("the", "and", "of"), and "abortion" (0.0186). Notably, "claims" appears in both true positives (as "claimed", "claiming", "claim") and false positives, suggesting the same framing language triggers correct and incorrect predictions depending on context. This context-sensitivity is a limitation of both models, though DA-RoBERTa-BABE-FT shows fewer instances.

\subsubsection{True Negatives}
Refer to Fig. \ref{fig:appendixB} below.
\begin{figure}[h]
    \centering
    \includegraphics[width=0.8\linewidth]{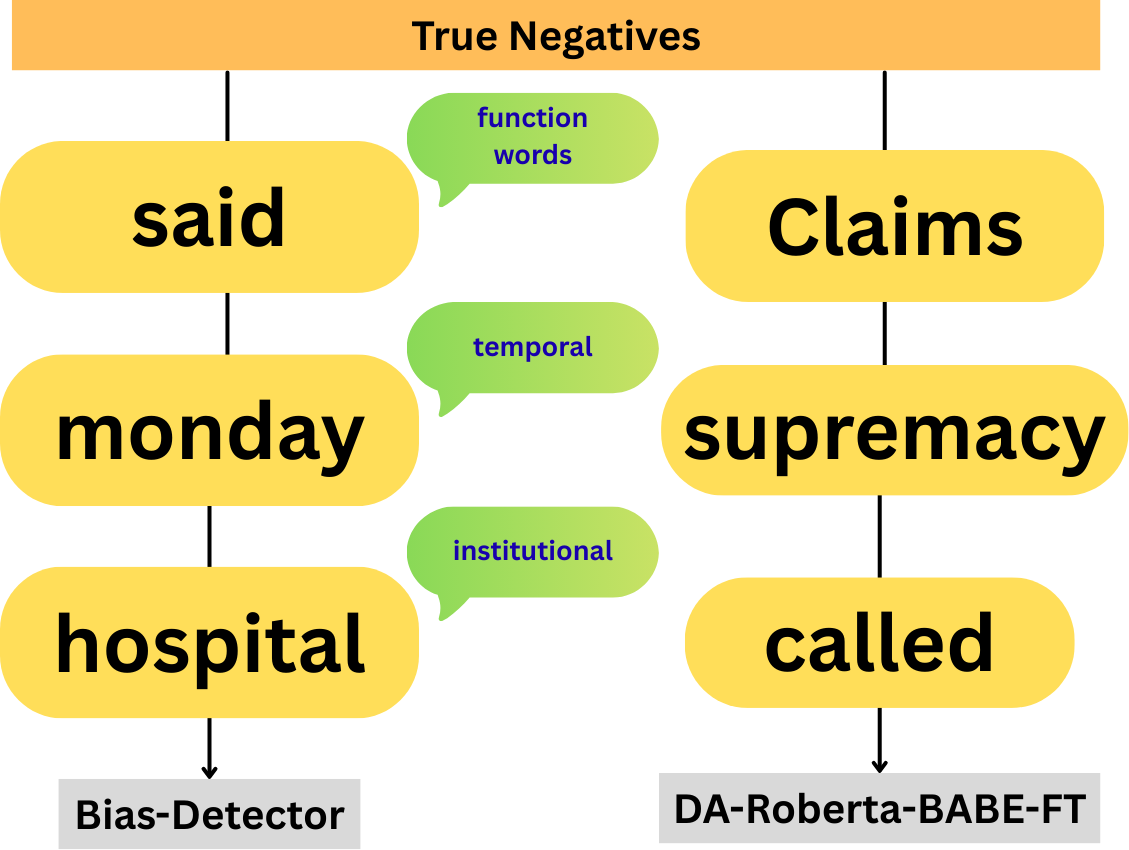}
    \caption{Examples of Biased Terms detected by both the models for the True Negatives.}
    \label{fig:appendixB}
\end{figure}

\end{document}